# Variational Multi-Modal Hypergraph Attention Network for Multi-Modal Relation Extraction

Qian Li
Beijing University of Posts and Telecommunications
Beijing, China
13240016260@163.com

Cheng Ji
Beihang University
Beijing, China
jicheng@act.buaa.edu.cn

Shu Guo
CNCERT of China
Beijing, China
guoshu@cert.org.cn

Yong Zhao
Beihang University
Beijing, China
zhaoyong@buaa.edu.cn

Qianren Mao
Zhongguancun Laboratory
Beijing, China
maoqr@zgclab.edu.cn

Shangguang Wang*
Beijing University of Posts and Telecommunications
Beijing, China
sgwang@bupt.edu.cn

Yuntao Wei
USTC University
Hefei, China
yuntaowei@mail.ustc.edu.cn

Jianxin Li
Beihang University
Beijing, China
lijx@act.buaa.edu.cn

## ABSTRACT

Multi-modal relation extraction (MMRE) is a challenging task that aims to identify relations between entities in text leveraging image information. Existing methods are limited by their neglect of the multiple entity pairs in one sentence sharing very similar contextual information (*i.e.*, the same text and image), resulting in increased difficulty in the MMRE task. To address this limitation, we propose the Variational Multi-Modal Hypergraph Attention Network (VM-HAN) for multi-modal relation extraction. Specifically, we first construct a multi-modal hypergraph for each sentence with the corresponding image, to establish different high-order intra-/inter-modal correlations for different entity pairs in each sentence. We further design the Variational Hypergraph Attention Networks (V-HAN) to obtain representational diversity among different entity pairs using Gaussian distribution and learn a better hypergraph structure via variational attention. VM-HAN achieves state-of-the-art performance on the MMRE task, outperforming existing methods in terms of accuracy and efficiency.

*Corresponding author

## 1 INTRODUCTION

Relation extraction (RE) is a crucial information extraction subtask that involves identifying the relationship between two entities [15, 30]. In recent years, there has been a growing trend towards multi-modal relation extraction (MMRE), which combines textual relations between entities with visual information. MMRE leverages multimedia content to provide additional knowledge that supports various cross-modal tasks, such as multi-modal knowledge graphs construction [25, 42] and visual question answering systems [1, 21].

Prior studies [3, 10, 14, 37, 38] have achieved remarkable success in multi-modal relation extraction by leveraging visual information, as visual content provides valuable evidence to supplement missing semantics. For example, one study [37] enriched text embeddings by introducing visual relations of objects in the image through an attention-based mechanism. Another approach, HVPNet [3], guided text representation learning using an object-level prefix and a multi-scale visual fusion mechanism. DGF-PT [14] adopts a prompt-based autoregressive encoder containing two types of prefix-tuning to build deeper correlations among entity pair, text, and image/objects. However, the above methods mainly focus on the relations between the objects and text and overlook the multiple entity pairs in one sentence that shares very similar contextual information (*i.e.*, the same text and image), leading to the failure to identify the different relations between different entity pairs.

In detail, a specific text and image often correspond to multiple entity pairs, which leads to different relations, as illustrated in Figure 1. These three entities can be combined into three entity pairs, corresponding to two different relations. Thus, for different entity pairs, the objects in the image have different importance. For example, the most important objects for the entity *Evan Massey* and the *Lori Blade* are the objects framed in blue, which are helpful to the task. It means that if the model can associate entities with multiple sets of objects, it can learn semantics that is more useful for relational classification. Furthermore, an entity may appear in multiple pairs, leading to diversity in the relations. For example,

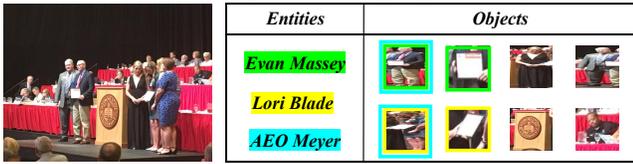

**Text: Evan Massey and Lori Blade presented with the prestigious AEO Meyer Harte Award.**

**Figure 1: An example of the MMRE task. The task is to predict the relation of given entity pairs for the specific text and image which contains multiple objects.**

the entity *Evan Massey* appears in two different entity pairs and some of the objects are helpful information for multiple entities, which enhances the difficulty of learning specific representations. The phenomenon shows that the proper association among entities and objects is vital for the MMRE task. In general, it is crucial to capture the correlations beyond pair-wise ones for entity pairs among different modalities and diverse representations of the same entities/relations. However, this still remains under-explored.

To overcome the above challenges, we propose the variational multi-modal hypergraph attention network (VM-HAN[1]) for MMRE. Unlike existing methods that rely on pre-defined or context features, our approach learns a joint representation of the multiple modalities by leveraging hypergraphs to capture complex, high-order correlations among different modalities. Specifically, we first construct a multi-modal hypergraph for MMRE, where each sentence and corresponding image (and the objects in the image) are represented as a hypergraph. Next, the model automatically learns the hypergraph's edges and weights to learn a better structure that captures the underlying associations between the entities and relations. To enhance the model's generalization ability and improve performance, we employ a method for variational modeling that transforms node representations into Gaussian distributions. It allows the model to handle the ambiguity or diversity in the relationships between entities and relations. By converting node representations into Gaussian distributions, the model can better capture the underlying distribution of relationships and generate more accurate predictions. The incorporation of hypergraphs and variational modeling allows our model to capture complex, high-order correlations between modalities and handle multiple different entities and relationships, making it a promising approach for real-world applications. Our proposed model achieves state-of-the-art performance on the MNRE task and outperforms existing methods in terms of both accuracy and efficiency.

- We technically design a novel MMRE framework to capture complex and high-order correlations among different modalities.
- We construct a multi-modal hypergraph capturing high-order correlations between different modalities. We also design Variational Hypergraph Attention Networks to handle the diversity of entities and relations.
- Experimental results on benchmark datasets demonstrate the effectiveness of our approach, outperforming state-of-the-art methods.

---

[1]The source code is available at https://anonymous.4open.science/r/VM-HAN-35E5/.

## 2 RELATED WORK

### 2.1 Multi-modal relation extraction (MMRE)

Multi-modal relation extraction (MMRE), a pivotal subdomain within the broader context of multi-modal information extraction in natural language processing (NLP), has garnered significant attention in recent years [17, 29, 35, 37]. MMRE aims to identify textual relations between two entities in a sentence by incorporating visual content [3, 37, 38], which compensates for insufficient semantics and aids in relation extraction [37, 38]. Recent works, such as MNRE [38] and HVPNet [37], have focused on MMRE technology and demonstrated the effectiveness of aggregating image information for relation extraction. MKGformer [2] and MOREformer [10] facilitate multi-level fusion between visual and textual information to process and integrate information from both modalities. DGF-PT [14] represents a recent advancement, employing a prompt-based autoregressive encoder that incorporates two distinct types of prefix-tuning to foster stronger connections between entity pairs, textual context, and image/objects. However, existing works ignore the multiple relations in different entity pairs in one sentence, which is caused by variations in entity pairs.

In the task of multi-modal relation extraction, it is observed that images can provide valuable information. However, the potential of utilizing image information in a distinct manner for different entity pairs has not been fully explored. This paper aims to address this gap by capturing high-order correlations among entity pairs and the associated image objects and integrating intra-modal and inter-modal correlations that are truly useful for each entity pair.

### 2.2 Hypergraph Network

Hypergraph networks have gained attention in various fields for modeling complex relationships among nodes [9, 27]. Unlike traditional graphs [26, 28, 40], hypergraphs allow hyperedges to connect more than two nodes, enabling the representation of higher-order relationships. Multimodal Hypergraph networks capture the intricate relationships between multimodal elements, have been successfully applied in tasks such as visual question answering [11], 3D object retrieval [6, 36], and image retrieval [33, 39]. In the realm of multimodal learning, [11] constructed a shared semantic space among different modalities by hypergraph and generated a joint representation by attentively integrating the modalities through a co-attention mechanism. MM3DOE [6] leveraged hypergraph structures to effectively model the complex correlations among 3D objects, even from unseen categories, by capturing the high-order dependencies within and across modalities. MKHG [33] presents a degree-free hypergraph solution that ingeniously addresses the challenges posed by heterogeneous data sources and modalities. By designing hyperlinks to connect diverse sub-semantics, this method enhances the ability to bridge and integrate information.

In hypergraph networks for the task, nodes represent entities and objects, and hyperedges represent relationships involving more than two nodes. This structure facilitates the modeling of multi-way interactions among nodes, capturing dependencies that are not adequately represented in traditional graphs [20, 34]. In this paper, we propose a novel approach that integrates hypergraph networks with multi-modal relation extraction to exploit high-order correlations among entity pairs and associated image objects.

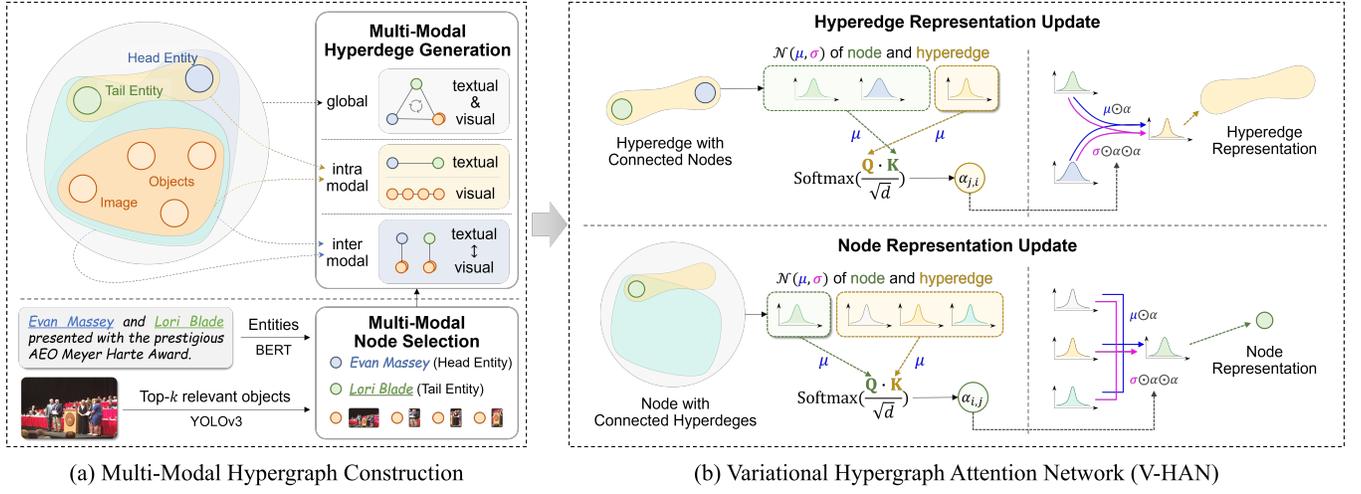

Figure 2: VM-HAN models text and corresponding images into a hypergraph for capturing high-order correlations and further learns entity pair representation under Gaussian distribution for robust nodes and hyperedges learning.

## 3 PRELIMINARIES

***Multi-Modal Relation Extraction Task (MMRE)***. Given a text $T = [w_1, w_2, \cdots, w_l]$ associated with an image $I$, and an entity pair $(h, t)$, an MMRE model takes $(h, t, T, I)$ as input and outputs a relation from a predefined set of relations $\mathcal{R} = \{r_1, r_2, \cdots, r_k, \text{none}\}$. The output relation contains visual clues, and a relation extractor is a trained model that computes a confidence score $p(r_i \mid h, t, T, I)$ for each relation $r_i \in \mathcal{R}$ to determine which $r_i$ is the relation between $h$ and $t$ in the given $T$ and $I$.

***Hypergraph***. A hypergraph is a specialized form of graph that differs from simple graphs by containing hyperedges. These hyperedges can connect two or more nodes and are often used to represent high-order correlations [8]. A hypergraph is typically defined as $\mathcal{G} = (\mathcal{V}, \mathcal{E})$, consisting of a node set $\mathcal{V} = \{v_1, v_2, \cdots, v_n\}$, a hyperedge set $\mathcal{E} = \{e_1, e_2, \cdots, e_m\}$. The hypergraph $\mathcal{G}$ can be represented by an incidence matrix $\mathbf{H} \in \{0, 1\}^{n \times m}$, where each entry is defined as:

$$\mathbf{H}_{i,j} = \begin{cases} 1, & \text{if } v_i \in e_j \\ 0, & \text{otherwise} \end{cases}.$$  (1)

In this way, each hyperedge $e_j$ connects all associated nodes $v_i$, indicating correlations among them.

## 4 FRAMEWORK

This section introduces our proposed variational multi-modal hypergraph attention network (VM-HAN) framework, as shown in Figure 2. We first construct a multi-modal hypergraph for each sentence $T$ with the corresponding image $I$, capturing high-order correlations between different modalities. To learn an optimal hypergraph, the model learns edge weights of hypergraph by a variational hypergraph attention network (V-HAN). The V-HAN employs a variational modeling approach that transforms node representations into Gaussian distributions to better capture underlying distribution of relationships. This allows model to handle ambiguity or diversity in relationships between entities and relations.

### 4.1 Multi-Modal Hypergraph Construction

MMRE involves extracting relationships between entities from both textual and visual modalities. To capture high-order correlations among different modalities, we propose a multi-modal hypergraph construction module. This module is designed to facilitate high-order relation understanding by exploiting pivotal correlations beyond pair-wise ones.

#### 4.1.1 Multi-Modal Node Selection.
The multi-modal hypergraph contains textual and visual nodes selected from the given text $T$ and image $I$. Specifically, the entities $h$ and $t$ in text are initialized as textual nodes. In addition, the visual nodes include the image itself and $k$ most relevant objects to the given entity pair $(h, t)$.

For textual nodes, we select the head and tail entities $(h, t)$ in text $T$ as nodes for the hypergraph. We use a text encoder (*e.g.*, BERT) to learn contextual representations of entities.

For visual nodes, we use the YOLOv3 [19] object detection algorithm to identify the top-$k$ most relevant objects to the entity pair in each image by semantic similarity, which are selected as additional visual information. Let $O = \{o_i\}_{i=1}^{k}$ denotes the objects. The image $I$ and objects $O$ are then transformed as visual nodes.

Finally, the constructed hypergraph contains two modalities of nodes, denoted as $\mathcal{V} = \{v_h, v_t\} \cup \{v_I, v_{o_1}, \cdots, v_{o_k}\}$, including textual nodes (head and tail entities) and visual nodes (origin image and top-$k$ most relevant objects).

#### 4.1.2 Multi-Modal Hyperedge Generation.
To establish high-order correlations beyond pair-wise ones, the nodes in multi-modal hypergraphs are connected through three types of hyperedges (*i.e.*, global, intra-modal, and inter-modal ones).

***Global Hyperedge***. To capture global correlations among all modalities, we first construct the global hyperedges connecting all nodes in $\mathcal{V}$. Formally, we define global hyperedge as follows:

$$e_{\text{global}} = \{v_h, v_t, v_I, v_{o_1}, \cdots, v_{o_k}\}. \tag{2}$$

Global hyperedge allows all nodes to propagate information to each other through the hyperedge and get representations that contain global semantics (*i.e.*, both textual and visual information).

***Intra-Modal Hyperedge.*** For relations within a single modality, we construct two intra-modal hyperedges for each hypergraph, one connecting the textual nodes (*i.e.*, head and tail entities) and the other connecting the visual nodes (*i.e.*, image and top-$k$ objects). The intra-modal hyperedges are defined as follows:

$$e_{\text{textual}} = \{v_h, v_t\}, e_{\text{visual}} = \{v_I, v_{o_1}, \cdots, v_{o_k}\}. \tag{3}$$

Intra-modal hyperedges focus on aggregating information within one single modality to obtain modal-specific semantics (*i.e.*, textual/visual correlations).

***Inter-Modal Hyperedge.*** In addition, to capture the relations across different modalities, we construct two different inter-modal hyperedges, one connecting the head entity with the visual nodes, and the other connecting the tail entity with the visual nodes. The inter-modal hyperedges are defined as follows:

$$\begin{aligned} e_{\text{head-visual}} &= \{v_h, v_I, v_{o_1}, \cdots, v_{o_k}\}, \\ e_{\text{tail-visual}} &= \{v_t, v_I, v_{o_1}, \cdots, v_{o_k}\}. \end{aligned} \tag{4}$$

Inter-modal hyperedges mainly help to learn the associations between modalities (*i.e.*, text ↔ image). $e_{\text{head-visual}}$ denotes the cross-modal hyperedge connecting the head entity with the image and top three objects, and $e_{\text{tail-visual}}$ denotes the cross-modal hyperedge connecting the tail entity with the image and top three objects.

Finally, the constructed multi-modal hypergraph contains three types of hyperedges, including global, intra-modal, and inter-modal hyperedges. To further study the strength of associations in the constructed multi-modal hypergraph, we next design an attention-based hypergraph encoder using variational modeling approaches.

The advantages of our proposed Multi-modal Hypergraph Construction module are twofold. First, by explicitly modeling the high-order correlations among different modalities, our module can capture more complex and fine-grained relations that might be overlooked by existing approaches. Second, by introducing hyperedges to connect multiple nodes simultaneously, our module can reduce the number of parameters needed to model the relationships among all nodes, which can reduce the risk of overfitting and improve the generalization performance of the model.

## 4.2 Variational Hypergraph Attention Network

To further improve the performance of MMRE on the constructed multi-modal hypergraph, we propose a Variational Hypergraph Attention Network (V-HAN), utilizing a variational approach to learn representations under Gaussian distributions.

### 4.2.1 Variational Hypergraph Representation.
The variational hypergraph representation is a crucial component of V-HAN that transfers the node features into latent distributions. It takes the textual and visual nodes as input and outputs representations modeled as Gaussian distributions.

To obtain the node representations, we model the feature of each node $v_i \in \mathcal{V}$ using a Gaussian distribution and compute the mean and variance of node features for transferring specific representation into Gaussian representation as follows:

$$\mathbf{x}_{i,\mu}^{(0)} = \mathbf{W}_\mu \cdot \mathbf{x}_i, \quad \mathbf{x}_{i,\sigma}^{(0)} = \mathbf{W}_\sigma \cdot \mathbf{x}_i. \tag{5}$$

Specifically, we model each node $v_i$ using a Gaussian distribution $\mathcal{N}(\mathbf{x}_{i,\mu}, \text{diag}(\mathbf{x}_{i,\sigma}))$, where $\mathbf{x}_{i,\mu}^{(0)}$ and $\mathbf{x}_{i,\sigma}^{(0)}$ are the initial representations of the mean and variance of the node feature distribution, respectively. $\mathbf{x}_i$ are the origin feature initialized in Section 4.1.1. $\mathbf{W}_\mu, \mathbf{W}_\sigma$ are learnable parameters. In this paper, we adopt the diagonal variance matrix, which is commonly used in studies [18, 41].

The variational representations preserve the original feature information in the mean vector and estimate the uncertainty of the original feature information in the variance vector. By modeling the node features as Gaussian distributions, this module provides a more informative and robust representation of each node, which helps for better multi-modal relation extraction performance.

### 4.2.2 Variational Hypergraph Attention.
Variational hypergraph attention is a module that updates node representations by propagating information through hyperedges.

To adaptively learn the influences under the structure of multi-modal hypergraphs, we deploy a multi-head attention mechanism to compute the weight between hyperedge $e_j$ and a node $v_i$ connected with it as follows:

$$\alpha_{j,i}^{(l)} = \underset{v \in e_j}{\text{Softmax}} \left( \frac{(\mathbf{W}_e^{(l)} \mathbf{e}_{j,\mu}^{(l)}) \cdot (\mathbf{W}_x^{(l)} \mathbf{x}_{i,\mu}^{(l)})^{\text{T}}}{\sqrt{d}} \right), \tag{6}$$

where $\mathbf{x}_{i,\mu}$ is the mean vector of node $v_i$ and $\mathbf{e}_{j,\mu}$ is the hyperedge representation updated below. $\mathbf{W}_x, \mathbf{W}_e$ are learnable transformation matrices and $d$ is the dimension of the $\mathbf{W}_e^{(l)} \mathbf{e}_{i,\mu}^{(l)}$ vector.

The variational hypergraph attention mechanism updates the Gaussian distribution of hyperedge $e_j$ by aggregating information from all nodes $v_i \in e_j$. Specifically, the mean and variance of hyperedge $e_j$ in the $(l+1)$-th layer of the updated distribution are computed using the following equations:

$$\begin{aligned} \mathbf{e}_{j,\mu}^{(l+1)} &= \sigma\left( \sum_{v_i \in e_j} \mathbf{x}_{i,\mu}^{(l)} \odot \alpha_{j,i}^{(l)} \cdot \mathbf{W}_{e,\mu}^{(l)} \right) + \mathbf{e}_{j,\mu}^{(l)}, \\ \mathbf{e}_{j,\sigma}^{(l+1)} &= \sigma\left( \sum_{v_i \in e_j} \mathbf{x}_{i,\sigma}^{(l)} \odot \alpha_{j,i}^{(l)} \odot \alpha_{j,i}^{(l)} \cdot \mathbf{W}_{e,\sigma}^{(l)} \right) + \mathbf{e}_{j,\sigma}^{(l)}, \end{aligned} \tag{7}$$

where $\mathbf{x}_{i,\mu}^{(l)}$ and $\mathbf{x}_{i,\sigma}^{(l)}$ are the mean and variance representation of node $v_i$ in $l$-th layer. In the variational framework, especially one that deals with Gaussian distributions, each node (or hyperedge) is represented by a mean ($\mu$) and a variance ($\sigma^2$). The mean represents the expected value of the feature, while the variance captures the uncertainty or spread of that feature's distribution. The double product in the updating of $\sigma$ (variance) reflects the need to account for the uncertainty in the attention weights $\alpha_{j,i}^{(l)}$ themselves, in addition to the uncertainty in the node or hyperedge representations. This is particularly important in a setting where attention mechanisms are used, as the attention weights determine how much influence one node has over another. The squaring (or double product) of $\sigma$ in this context ensures that the variance is scaled appropriately, reflecting the compounded uncertainty that arises from both the attention mechanism and the variational representations.

For node representation updating, similar to the hyperedges, we compute the weight between node $v_i$ and a hyperedge $e_j$ connected with it as follows:

$$\alpha_{i,j}^{(l)} = \underset{e \in v_i}{\text{Softmax}} \left( \frac{(\mathbf{W}_x^{(l)} x_{i,\mu}^{(l)}) \cdot (\mathbf{W}_e^{(l)} e_{j,\mu}^{(l)})^\mathrm{T}}{\sqrt{d}} \right). \quad (8)$$

The updated Gaussian distribution of node $v_i$ is computed by aggregating information from all hyperedges connected to it. Specifically, the mean and variance of node $v_i$ in the $(l+1)$-th layer of the updated distribution are computed using the following equations:

$$\begin{aligned}
x_{i,\mu}^{(l+1)} &= \sigma \left( \sum_{e_j \in v_i} e_{j,\mu}^{(l)} \odot \alpha_{i,j}^{(l)} \cdot \mathbf{W}_{x,\mu}^{(l)} \right) + x_{i,\mu}^{(l)}, \\
x_{i,\sigma}^{(l+1)} &= \sigma \left( \sum_{e_j \in v_i} e_{j,\sigma}^{(l)} \odot \alpha_{i,j}^{(l)} \odot \alpha_{i,j}^{(l)} \cdot \mathbf{W}_{x,\sigma}^{(l)} \right) + x_{i,\sigma}^{(l)},
\end{aligned} \quad (9)$$

where $e_{j,\mu}^{(l)}$ and $e_{j,\sigma}^{(l)}$ are the mean and variance representation of hyperedge $e_j$ in $l$-th layer.

Overall, V-HAN can capture higher-order correlations among the different modalities and improve its ability to model complex relationships by iteratively updating the attention between nodes and hyperedges. Furthermore, it can learn informative and robust representations for the nodes and hyperedges by modeling them as Gaussian distributions. Together with the carefully-designed joint optimization objective function in the next section, the proposed VM-HAN improves the performance of MMRE task.

### 4.3 Joint Optimization Objectives

In our proposed VM-HAN, we design a joint loss function to evaluate completion under variational representation. The loss function includes three parts, relation classification loss, reconstruction loss, and KL divergence loss.

The relation classification loss is used to measure the classification error between the predicted relation and the ground truth relation. The relation classification loss is defined as follows:

$$\mathcal{L}_c = -\log p(r | x_{h,\mu}, x_{h,\sigma}, x_{t,\mu}, x_{t,\sigma}), \quad (10)$$

where **r** is the relation between the head entity $v_h$ and the tail entity $v_t$, and $(x_{h,\mu}, x_{h,\sigma}), (x_{t,\mu}, x_{t,\sigma})$ are the representations of the two entities. $p(r | x_{h,\mu}, x_{h,\sigma}, x_{t,\mu}, x_{t,\sigma})$ is derived by concatenating four vectors followed by a classification layer.

To ensure that the predicted node representations are close to the actual representations and that the predicted distributions are similar to a standard Gaussian, we introduce reconstruction loss and KL divergence loss. The reconstruction loss $\mathcal{L}_\text{rec}$ measures the difference between the predicted node representations and the actual node representations. It is defined as the mean squared error (MSE) between the predicted mean vectors $x_{i,\mu}$ and the true mean vectors $x_{i,h}$, and the MSE between the predicted variance vectors $x_{i,\sigma}$ and the true variance vectors $x_{i,\tau}$:

$$\mathcal{L}_\text{rec} = \frac{1}{n} \sum_{i=1}^n \left( |x_{i,h} - x_{i,\mu}|^2 + |x_{i,\tau} - x_{i,\sigma}|^2 \right), \quad (11)$$

where $n$ is the number of nodes in the multi-modal hypergraph. The KL divergence loss $\mathcal{L}_\text{KL}$ measures the difference between the predicted Gaussian distributions and a standard Gaussian distribution. It is defined as:

$$\mathcal{L}_\text{KL} = \frac{1}{n} \sum_{i=1}^n D_\text{KL}(\mathcal{N}(x_{i,\mu}, \text{diag}(x_{i,\sigma})) | \mathcal{N}(0, \mathbf{I})), \quad (12)$$

where $D_\text{KL}$ is the Kullback-Leibler divergence and $\mathcal{N}(\mu, \Sigma)$ is the multivariate Gaussian distribution with mean $\mu$ and covariance matrix $\Sigma$. The KL loss is to regularize the variational distribution, thereby improving the generalization ability of the model. The KL loss in the proposed model serves as a regularization term to encourage the learned latent distribution to be close to a prior distribution. It helps prevent overfitting and improves the generalization ability of the model.

The overall loss function of VM-HAN is a weighted sum of the three loss terms:

$$\mathcal{L} = \lambda_c \mathcal{L}_c + \lambda_\text{rec} \mathcal{L}_\text{rec} + \lambda_\text{KL} \mathcal{L}_\text{KL}, \quad (13)$$

where $\lambda_c, \lambda_\text{rec}$, and $\lambda_\text{KL}$ are hyperparameters that control the relative importance of each loss term.

## 5 EXPERIMENT

### 5.1 Dataset and Evaluation Metric

We evaluate our model on the two following MMRE datasets:

**MNRE.** We conduct experiments in MNRE dataset [38], crawling data from Twitter[2]. MNRE includes 15,484 samples and 9,201 images. It contains 23 relation categories. As previous works [37] recommended, MNRE is splitted into 12,247 training samples, 1,624 development samples and 1,614 testing samples.

**MORE.** We further extend our research using MORE dataset [10]. The MORE dataset is comprised of 21 unique types of relations and encompasses 20,264 annotated multimodal relational facts across 3,559 text-image pairs, positioning it as one of the most extensive datasets for multimodal relation extraction across various dimensions. Additionally, it incorporates 13,520 visual objects, averaging 3.8 objects per image, a vital component previously absent in earlier datasets. The dataset is divided into training, development, and testing segments, containing 15,486, 1,742, and 3,036 facts, respectively.

We report official Accuracy (Acc.), Precision (Prec.), Recall, and F1 metrics for evaluation.

### 5.2 Comparision Methods

We compare our method with three text-based RE models, eight BERT-based MMRE models, and three graph neural networks for the MMRE task.

We compare our method with three text-based RE models : (1) **Glove+CNN** [32] is a CNN-based model with an additional position embeddings. (2) **PCNN** [31] is relation extraction method utilizing external knowledge graph with a distantly supervision manner. (3) **Matching the Blanks (MTB)** [22] is a BERT based relation extraction method without visual features.

We further consider another group of previous approaches for MMRE to integrate visual information: (4) **BERT+SG** [4] concatenates BERT representations with visual contents, which is obtained by the pre-trained scene graph tool [23]. (5) **BERT+SG+Att** adopts an attention mechanism to compute the relevance between textual and visual features. (6) **VisualBERT** [13] is a single-stream encoder, learning cross-modal correlation in a model. (7) **MEGA** [37] considers the relevance from structural and semantic perspective with graph alignment. (8) **HVPNet** [3] introduces an object-level prefix with a dynamic gated aggregation strategy to enhance the correlation between all objects and text. (9) **DGF-PT** [14] adopts a prompt-based autoregressive encoder containing two types of prefix-tuning to build deeper correlations among entity pair, text, and image/objects. (10) **MKGformer** [2] introduces an innovative M-Encoder module that facilitates a multi-level fusion between visual and textual Transformers, enhancing the model's capability to process and integrate information from both modalities. (11) **MOREformer** [10] employs a sophisticated fusion of attribute-aware textual encoding, depth-aware visual encoding, and position-fused multimodal encoding, thereby augmenting its proficiency in multi-object disambiguation.

---
[2]https://archive.org/details/twitterstream

Table 1: Main experiments on MNRE and MORE. The best results are highlighted in bold, "–" means results are not available, and the underlined values are the second-best result. "↑" means the increase compared to the underlined values.

| Model Type | Model Name | MNRE dataset | | | | MORE dataset | | | |
| --- | --- | --- | --- | --- | --- | --- | --- | --- | --- |
| | | Acc. (%) | Prec. (%) | Recall (%) | F1 (%) | Acc. (%) | Prec. (%) | Recall (%) | F1 (%) |
| Text-based RE | Glove+CNN [32] | 70.32 | 57.81 | 46.25 | 51.39 | 60.23 | 27.87 | 31.65 | 29.64 |
| | PCNN [31] | 72.67 | 62.85 | 49.69 | 55.49 | 59.35 | 28.02 | 38.31 | 32.37 |
| | MTB [22] | 72.73 | 64.46 | 57.81 | 60.96 | 60.19 | 29.40 | 40.37 | 34.02 |
| BERT-based MMRE | BERT+SG [4] | 74.09 | 62.95 | 62.65 | 62.80 | 61.79 | 29.61 | 41.27 | 34.48 |
| | BERT+SG+Att. [4] | 74.59 | 60.97 | 66.56 | 63.64 | 63.74 | 31.10 | 39.28 | 34.71 |
| | VisualBERT [13] | - | 57.15 | 59.45 | 58.30 | 82.84 | 58.18 | 61.22 | 59.66 |
| | MEGA [37] | 76.15 | 64.51 | 68.44 | 66.41 | 65.97 | 33.30 | 38.53 | 35.72 |
| | HVPNet [3] | 90.95 | 83.64 | 80.78 | 81.85 | 72.40 | 61.47 | 65.26 | 63.31 |
| | MKGformer [2] | 83.36 | 82.40 | 81.73 | 82.06 | 80.17 | 55.76 | 53.74 | 54.73 |
| | MOREformer [10] | 82.67 | 82.19 | 82.35 | 82.27 | 83.50 | 62.18 | 63.34 | 62.75 |
| | DGF-PT [14] | 84.25 | 84.35 | 83.83 | 84.47 | 82.35 | 60.51 | 62.82 | 61.64 |
| Graph-based MMRE | GCN [12] | 73.64 | 63.70 | 67.41 | 65.50 | 78.28 | 59.57 | 59.10 | 59.33 |
| | GAT [24] | 78.50 | 67.26 | 70.37 | 68.78 | 79.26 | 58.52 | 60.49 | 59.49 |
| | HGNN [7] | 83.21 | 71.92 | 73.94 | 72.92 | 81.94 | 62.27 | 61.05 | 61.65 |
| | VM-HAN (Ours) | **93.57** (↑2.62) | **85.76** (↑1.41) | **84.69** (↑0.86) | **85.22** (↑0.75) | **85.57** (↑2.07) | **64.76** (↑2.49) | **66.69** (↑1.33) | **65.71** (↑2.40) |

We further train on three graph neural networks for MMRE: (12) **GCN** [12] uses graph convolutional layers. In our case, we use GCN to learn node embeddings by aggregating information from its neighboring nodes in the graph. (13) **GAT** [24] uses attention mechanisms to learn node representations. (14) **HGNN** [7] is a hypergraph graph neural network. In HGNN, nodes are associated with hyperedges, and the attention mechanism is applied over hyperedges instead of edges in a traditional graph.

### 5.3 Implementation Details

We used PyTorch[3] as a deep learning framework to develop the MMRE. All experiments were conducted on a server with one GPU (Tesla V100). The BERT version is bert-base-uncased in hugging-face[4] for text initialization and set the dimension at 768. The VGG version is VGG16[5] and Faster R-CNN [5] for image initialization and set the dimension of visual objects features at 4096. The maximum length of text is 128 and objects of each image is 3. During training, we optimize the model using cross-entropy loss. The learning rate is 2e-5, and batch size is 16, and dropout rate is 0.6. We use AdamW [16] to optimize parameters with a weight decay of 0.01 and deploy grid search for optimal hyperparameters. The node embeddings are learned using a GCN, and the hypergraph attention mechanism is used to aggregate information from different nodes and hyperedges. To ensure fairness, all baseline models use the same data set partitioning. All hyper-parameter settings are tuned on the validation data by the grid search with 5 trials.

### 5.4 Main Results

To verify the effectiveness of our model, the latest Text-based, BERT-based, GNN-based relation extraction models are selected as the baseline models. We perform relation extraction on MNRE, MORE, and MFewRel datasets and report the average results with the standard deviation across the 10-folds cross-validation in Table 1. Part of the results are from the original papers [3, 10, 14, 37]. From the table, we can observe that: 1) Our model outperforms all baselines, in terms of two MMRE datasets. It demonstrates our model incorporates more multi-modal knowledge, achieving reliable performance. 2) Our model outperforms all baselines, in terms of four metrics. Especially, our model improves at least 1.35% on F1 and 2.62% on Acc., respectively. It demonstrates our model incorporates more multi-modal knowledge, achieving reliable performance. 3) Compared to text-based RE models, our model achieves better performance, indicating its ability to effectively leverage visual information to enhance model performance. 4) Compared to BERT-based MMRE models, our model performs better, highlighting its ability to capture structural features through hypergraph learning. 5) Our model also outperforms graph neural networks, indicating that our hypergraph learning method can capture more high-order correlations and variational modeling, leading to better performance. 5) It should be noted that the improvement of VM-HAN is more significant on MORE dataset. It improves at least 2.07 points on the for evaluation metrics. The major reason is that MORE contains abundant objects, which helps to capture more cross-modal fine-grained relations. Taken together, VM-HAN achieved peak performance across all datasets, demonstrating the need to model multimodal hypergraph and variational representation. All observations demonstrate that our model is effective in leveraging multi-modal information and capturing complex correlations.

---
[3] https://pytorch.org/
[4] https://github.com/huggingface/transformers
[5] https://github.com/machrisaa/tensorflow-vgg

Table 2: Ablation study on MNRE and MORE. "w/o" means removing the corresponding module from the complete model. "repl." means replacing the corresponding module. The "↓" means the performance degradation compared to the VM-HAN.

| Variants | MNRE dataset | | | | | MORE dataset | | | | |
|---|---|---|---|---|---|---|---|---|---|---|
| | Acc. (%) | Prec. (%) | Recall (%) | F1 (%) | △ Avg (%) | Acc. (%) | Prec. (%) | Recall (%) | F1 (%) | △ Avg (%) |
| **VM-HAN (Ours)** | **93.57** | **85.76** | **84.69** | **85.22** | - | **85.57** | **64.76** | **66.69** | **65.71** | - |
| w/o Multi-Modal Hypergraph | 90.36 | 81.82 | 79.55 | 80.67 | ↓ 4.21 | 82.45 | 61.28 | 62.01 | 61.64 | ↓ 3.84 |
| w/o Variational Representation | 91.74 | 81.53 | 79.81 | 80.66 | ↓ 3.88 | 82.71 | 61.30 | 63.94 | 62.59 | ↓ 3.05 |
| w/o V-HAN | 88.29 | 80.41 | 77.60 | 78.98 | ↓ 5.99 | 80.60 | 59.52 | 60.39 | 59.95 | ↓ 5.57 |
| w/o KL Loss | 92.68 | 83.85 | 81.97 | 82.90 | ↓ 1.96 | 83.53 | 62.28 | 63.42 | 62.84 | ↓ 2.67 |
| w/o Global Hypergraph | 92.43 | 84.53 | 81.28 | 82.87 | ↓ 2.03 | 83.25 | 62.73 | 64.26 | 63.49 | ↓ 2.25 |
| w/o Intra-modal Hypergraph | 92.36 | 83.76 | 80.64 | 82.17 | ↓ 2.58 | 83.29 | 62.63 | 64.72 | 63.66 | ↓ 2.11 |
| w/o Inter-modal Hypergraph | 92.80 | 83.69 | 80.25 | 81.93 | ↓ 2.64 | 83.39 | 62.05 | 64.28 | 63.15 | ↓ 2.47 |
| repl. HGNN | 87.72 | 80.38 | 79.36 | 79.87 | ↓ 5.48 | 80.48 | 60.52 | 62.62 | 61.55 | ↓ 4.39 |
| repl. Variational GCN | 88.49 | 80.64 | 80.19 | 80.41 | ↓ 4.88 | 81.53 | 59.18 | 61.47 | 60.30 | ↓ 5.06 |
| repl. GCN | 89.57 | 79.20 | 80.53 | 79.86 | ↓ 5.02 | 81.27 | 60.46 | 62.39 | 61.41 | ↓ 4.30 |
| repl. GAT | 89.61 | 80.75 | 80.22 | 80.48 | ↓ 4.55 | 81.72 | 60.29 | 62.04 | 61.15 | ↓ 4.38 |

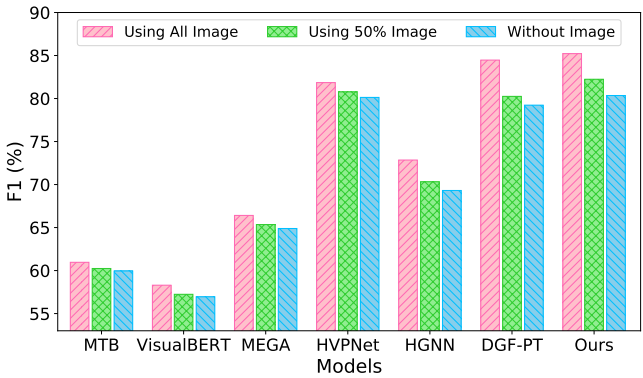

Figure 3: Different proportions of visual information on MNRE. "Without Image" means deleting all images. "Using 50% Image" means randomly deleting 50% images.

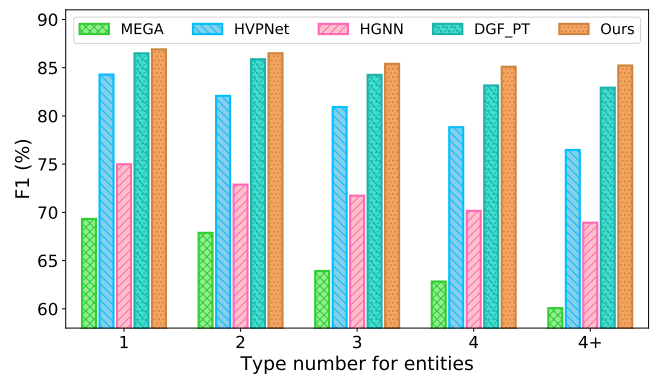

Figure 4: Impact of differences in sample number on MNRE. It means the performance (F1) when an entity belongs to one or multiple entity types.

### 5.5 Ablation Study

To investigate the effectiveness of each module in VM-HAN, we conduct variant experiments, showcasing the results in Table 2. The "↓" means the value of performance degradation compared to the VM-HAN. From the table, we can observe that: 1) For the all three multimodal relation extraction datasets, the kernel modules of our model improve the performance. It demonstrates that the multimodal hypergraph and variational representation are effective for the extraction task. 2) The impact of the multi-modal hypergraph tends to be more significant than other modules. We believe this is because the multi-modal hypergraph facilitates high-order relations, thus providing more clues. 3) The replacement of variational representations with specific representations resulted in a decline in performance. This indicates that our full model can better capture the underlying distribution of relationships and generate more accurate predictions. 4) By removing any one of the hypergraphs, the performance basically decreased. It demonstrates that each hypergraph is helpful for extraction and captures different high-order relations. 5) By replacing the Variational Hypergraph Network with other GNN models (such as GCN) resulted in decreased performance, suggesting that our model excels in handling situations with significant ambiguity or diversity in the relationships between entities and relations.

### 5.6 Effect of Visual Information

To validate the effectiveness of visual information in our proposed VM-HAN model, we conduct an experiment where we compare the performance of our model with and without image features, as shown in Figure 3 and more experiments on Appendix A. Specifically, we train two versions of the VM-HAN model: one that uses both textual and visual information, and another that only uses textual information. We evaluate the performance of both models on the same test set and report their precision, recall, and F1 scores. The results of the experiment show that the VM-HAN model with visual features outperforms the model without visual features by a significant margin, achieving higher precision, recall, and F1 scores. This indicates that incorporating visual information can enhance the ability of the model to capture high-order correlations between different modalities and improve the accuracy of multi-modal relation extraction. The experiment also demonstrates the effectiveness of the proposed VM-HAN model in handling multi-modal data and the importance of considering visual information in MMRE.

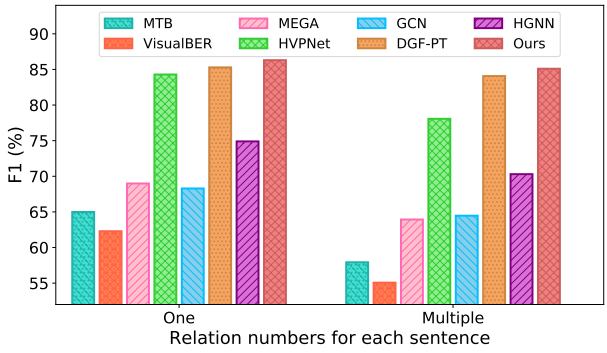

Figure 5: Impact of relation numbers for each sentence.

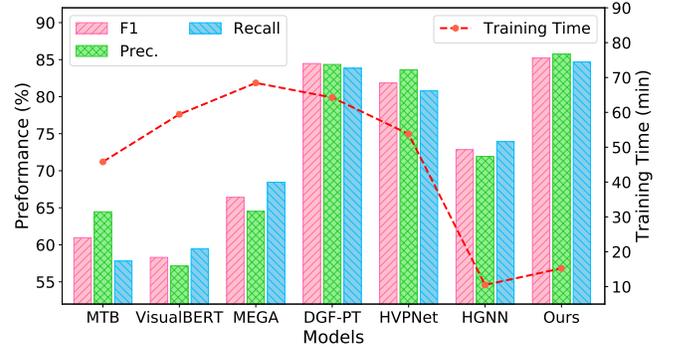

Figure 6: Performance and efficiency of models.

### 5.7 Discussions for V-HAN

We investigated the impact of multi-semantic entities on relation extraction. One situation is when an entity appears multiple times in different categories. We conducted experiments to compare the performance of our proposed method with a baseline method on different numbers of entity repetitions, as shown in Figure 4 and more experiments on Appendix B. Our method outperformed the baseline in both single and multiple entity repetition scenarios, indicating that our method effectively models the diversity of multi-semantic entities and learns their representations to improve extraction.

Another situation is when an entity has different meanings in different entity pairs within the same sentence, resulting in different relation types. We compared the performance of our method with the baselines on different entity pair numbers in a sentence, as shown in Figure 5 and more experiments on Appendix B. Our method achieved state-of-the-art performance on both single and multiple entity pairs scenarios, indicating that our proposed hypergraph learning model effectively utilizes inter-modal associations of different entity pairs to capture differences.

### 5.8 Efficiency

We conducted experiments to compare the efficiency of our proposed hypergraph-based approach for relation extraction with other baseline models, as illustrated in Figure 6. Our model demonstrates significantly reduced training time compared to BERT-based models, which are widely used but computationally intensive. Additionally, unlike graph neural network methods such as GCN, our approach constructs hypergraphs specifically for entity pairs and visual information. This focused construction allows us to learn high-order features within and across modalities more efficiently. By leveraging hypergraph structures tailored to the task at hand, our method streamlines the learning process, leading to faster convergence during training. Consequently, our approach not only achieves state-of-the-art accuracy in relation extraction tasks but also notably improves efficiency in terms of training time.

### 5.9 Case Study

To illustrate our model can effectively identify useful visual information, we provide an example involving various entity pairs. As shown in Fig. 7, the helpful information varies depending on the entity pair. From the figure, we can observe that: 1) Our model achieves superior performance across different entity pairs. It demonstrates its capability to effectively capture and model the correlations among entity pairs, images, and their associated objects. By leveraging hypergraph-based representations, our model can discern relevant visual cues for each entity pair. 2) When the sentence contains multiple entity pairs and the corresponding entity types are different, our method can predict the relation. It demonstrates that our model can effectively navigate the complexities of multi-entity relations by capturing diverse and high-order correlations in the hypergraph in handling complex associations between entity pairs.

## 6 CONCLUSION

The incorporation of hypergraphs and variational modeling allows our model to capture complex, high-order correlations between modalities and handle polysemous entities and relations. The model automatically learns the edge weight of the hypergraph, allowing it to capture the underlying associations between entities and relations. Moreover, the variational modeling approach employed by our Variational Hypergraph Attention Networks enables the model to handle multiple semantics of entities. Our proposed VM-HAN model achieves state-of-the-art performance on the MMRE task, outperforming existing methods.

Variational Multi-Modal Hypergraph Attention Network for Multi-Modal Relation Extraction ACM MM, 2024, Melbourne, Australia# REFERENCES

[1] Stanislaw Antol, Aishwarya Agrawal, Jiasen Lu, Margaret Mitchell, Dhruv Batra, C Lawrence Zitnick, and Devi Parikh. 2015. Vqa: Visual question answering. In *ICCV*. 2425–2433.

[2] Xiang Chen, Ningyu Zhang, Lei Li, Shumin Deng, Chuanqi Tan, Changliang Xu, Fei Huang, Luo Si, and Huajun Chen. 2022. Hybrid Transformer with Multi-level Fusion for Multimodal Knowledge Graph Completion. In *SIGIR '22: The 45th International ACM SIGIR Conference on Research and Development in Information Retrieval, Madrid, Spain, July 11 - 15, 2022*, Enrique Amigó, Pablo Castells, Julio Gonzalo, Ben Carterette, J. Shane Culpepper, and Gabriella Kazai (Eds.). ACM, 904–915. https://doi.org/10.1145/3477495.3531992

[3] Xiang Chen, Ningyu Zhang, Lei Li, Yunzhi Yao, Shumin Deng, Chuanqi Tan, Fei Huang, Luo Si, and Huajun Chen. 2022. Good Visual Guidance Make A Better Extractor: Hierarchical Visual Prefix for Multimodal Entity and Relation Extraction. In *NAACL*. 1607–1618.

[4] Jacob Devlin, Ming-Wei Chang, Kenton Lee, and Kristina Toutanova. 2019. BERT: Pre-training of Deep Bidirectional Transformers for Language Understanding. In *NAACL-HLT*. Association for Computational Linguistics, 4171–4186. https://doi.org/10.18653/v1/n19-1423

[5] RCNN Faster. 2015. Towards real-time object detection with region proposal networks. *Advances in neural information processing systems* 9199, 10.5555 (2015), 2969239–2969250.

[6] Yifan Feng, Shuyi Ji, Yu-Shen Liu, Shaoyi Du, Qionghai Dai, and Yue Gao. 2024. Hypergraph-Based Multi-Modal Representation for Open-Set 3D Object Retrieval. *IEEE Trans. Pattern Anal. Mach. Intell.* 46, 4 (2024), 2206–2223. https://doi.org/10.1109/TPAMI.2023.3332768

[7] Yifan Feng, Haoxuan You, Zizhao Zhang, Rongrong Ji, and Yue Gao. 2018. Hypergraph Neural Networks. *CoRR* abs/1809.09401 (2018). arXiv:1809.09401 http://arxiv.org/abs/1809.09401

[8] Yifan Feng, Haoxuan You, Zizhao Zhang, Rongrong Ji, and Yue Gao. 2019. Hypergraph neural networks. In *Proceedings of the AAAI conference on artificial intelligence*, Vol. 33. 3558–3565.

[9] Yue Gao, Yifan Feng, Shuyi Ji, and Rongrong Ji. 2023. HGNN$^+$: General Hypergraph Neural Networks. *IEEE Trans. Pattern Anal. Mach. Intell.* 45, 3 (2023), 3181–3199. https://doi.org/10.1109/TPAMI.2022.3182052

[10] Liang He, Hongke Wang, Yongchang Cao, Zhen Wu, Jianbing Zhang, and Xinyu Dai. 2023. MORE: A Multimodal Object-Entity Relation Extraction Dataset with a Benchmark Evaluation. In *Proceedings of the 31st ACM International Conference on Multimedia, MM 2023, Ottawa, ON, Canada, 29 October 2023- 3 November 2023*, Abdulmotaleb El-Saddik, Tao Mei, Rita Cucchiara, Marco Bertini, Diana Patricia Tobon Vallejo, Pradeep K. Atrey, and M. Shamim Hossain (Eds.). ACM, 4564–4573. https://doi.org/10.1145/3581783.3612209

[11] Eun-Sol Kim, Woo-Young Kang, Kyoung-Woon On, Yu-Jung Heo, and Byoung-Tak Zhang. 2020. Hypergraph Attention Networks for Multimodal Learning. In *2020 IEEE/CVF Conference on Computer Vision and Pattern Recognition, CVPR 2020, Seattle, WA, USA, June 13-19, 2020*. Computer Vision Foundation / IEEE, 14569–14578. https://doi.org/10.1109/CVPR42600.2020.01459

[12] Thomas N. Kipf and Max Welling. 2016. Semi-Supervised Classification with Graph Convolutional Networks. *CoRR* abs/1609.02907 (2016). arXiv:1609.02907 http://arxiv.org/abs/1609.02907

[13] Liunian Harold Li, Mark Yatskar, Da Yin, Cho-Jui Hsieh, and Kai-Wei Chang. 2019. VisualBERT: A Simple and Performant Baseline for Vision and Language. *CoRR* abs/1908.03557 (2019).

[14] Qian Li, Shu Guo, Cheng Ji, Xutan Peng, Shiyao Cui, Jianxin Li, and Lihong Wang. 2023. Dual-Gated Fusion with Prefix-Tuning for Multi-Modal Relation Extraction. In *Findings of the Association for Computational Linguistics: ACL 2023, Toronto, Canada, July 9-14, 2023*, Anna Rogers, Jordan L. Boyd-Graber, and Naoaki Okazaki (Eds.). Association for Computational Linguistics, 8982–8994. https://doi.org/10.18653/V1/2023.FINDINGS-ACL.572

[15] Haozhuang Liu, Ziran Li, Dongming Sheng, Hai-Tao Zheng, and Ying Shen. 2021. Multi-Entity Collaborative Relation Extraction. In *ICASSP 2021 - 2021 IEEE International Conference on Acoustics, Speech and Signal Processing (ICASSP)*. 7678–7682. https://doi.org/10.1109/ICASSP39728.2021.9413673

[16] Ilya Loshchilov and Frank Hutter. 2019. Decoupled Weight Decay Regularization. In *ICLR*. OpenReview.net. https://openreview.net/forum?id=Bkg6RiCqY7

[17] Yaojie Lu, Qing Liu, Dai Dai, Xinyan Xiao, Hongyu Lin, Xianpei Han, Le Sun, and Hua Wu. 2022. Unified Structure Generation for Universal Information Extraction. In *ACL*. Association for Computational Linguistics, 5755–5772. https://doi.org/10.18653/v1/2022.acl-long.395

[18] Valentin V Petrov. 2022. Sums of independent random variables. In *Sums of Independent Random Variables*. De Gruyter.

[19] Joseph Redmon and Ali Farhadi. 2018. Yolov3: An incremental improvement. *arXiv preprint arXiv:1804.02767* (2018).

[20] Franco Scarselli, Marco Gori, Ah Chung Tsoi, Markus Hagenbuchner, and Gabriele Monfardini. 2008. The graph neural network model. *IEEE transactions on neural networks* 20, 1 (2008), 61–80.

[21] Kevin J Shih, Saurabh Singh, and Derek Hoiem. 2016. Where to look: Focus regions for visual question answering. In *CVPR*. 4613–4621.

[22] Livio Baldini Soares, Nicholas FitzGerald, Jeffrey Ling, and Tom Kwiatkowski. 2019. Matching the Blanks: Distributional Similarity for Relation Learning. In *ACL*. Association for Computational Linguistics, 2895–2905. https://doi.org/10.18653/v1/p19-1279

[23] Kaihua Tang, Yulei Niu, Jianqiang Huang, Jiaxin Shi, and Hanwang Zhang. 2020. Unbiased Scene Graph Generation From Biased Training. In *CVPR*. Computer Vision Foundation / IEEE, 3713–3722. https://doi.org/10.1109/CVPR42600.2020.00377

[24] Petar Velickovic, Guillem Cucurull, Arantxa Casanova, Adriana Romero, Pietro Liò, and Yoshua Bengio. 2018. Graph Attention Networks. In *6th International Conference on Learning Representations, ICLR 2018, Vancouver, BC, Canada, April 30 - May 3, 2018, Conference Track Proceedings*. OpenReview.net. https://openreview.net/forum?id=rJXMpikCZ

[25] Meng Wang, Guilin Qi, HaoFen Wang, and Qiushuo Zheng. 2019. Richpedia: a comprehensive multi-modal knowledge graph. In *Joint International Semantic Technology Conference*. Springer, 130–145.

[26] Min Wang, Hao Yang, and Qing Cheng. 2022. GCL: Graph Calibration Loss for Trustworthy Graph Neural Network. In *MM '22: The 30th ACM International Conference on Multimedia, Lisboa, Portugal, October 10 - 14, 2022*, João Magalhães, Alberto Del Bimbo, Shin'ichi Satoh, Nicu Sebe, Xavier Alameda-Pineda, Qin Jin, Vincent Oria, and Laura Toni (Eds.). ACM, 988–996. https://doi.org/10.1145/3503161.3548423

[27] Xiangping Wu, Qingcai Chen, Wei Li, Yulun Xiao, and Baotian Hu. 2020. AdaHGNN: Adaptive Hypergraph Neural Networks for Multi-Label Image Classification. In *MM '20: The 28th ACM International Conference on Multimedia, Virtual Event / Seattle, WA, USA, October 12-16, 2020*, Chang Wen Chen, Rita Cucchiara, Xian-Sheng Hua, Guo-Jun Qi, Elisa Ricci, Zhengyou Zhang, and Roger Zimmermann (Eds.). ACM, 284–293. https://doi.org/10.1145/3394171.3414046

[28] Zonghan Wu, Shirui Pan, Fengwen Chen, Guodong Long, Chengqi Zhang, and S Yu Philip. 2020. A comprehensive survey on graph neural networks. *IEEE transactions on neural networks and learning systems* 32, 1 (2020), 4–24.

[29] Zhiwei Wu, Changmeng Zheng, Yi Cai, Junying Chen, Ho-fung Leung, and Qing Li. 2020. Multimodal Representation with Embedded Visual Guiding Objects for Named Entity Recognition in Social Media Posts. In *ACM MM*. ACM, 1038–1046. https://doi.org/10.1145/3394171.3413650

[30] Fuzhao Xue, Aixin Sun, Hao Zhang, Jinjie Ni, and Eng-Siong Chng. 2022. An Embarrassingly Simple Model for Dialogue Relation Extraction. In *ICASSP 2022 - 2022 IEEE International Conference on Acoustics, Speech and Signal Processing (ICASSP)*. 6707–6711. https://doi.org/10.1109/ICASSP43922.2022.9747486

[31] Daojian Zeng, Kang Liu, Yubo Chen, and Jun Zhao. 2015. Distant Supervision for Relation Extraction via Piecewise Convolutional Neural Networks. In *EMNLP*. The Association for Computational Linguistics, 1753–1762. https://doi.org/10.18653/v1/d15-1203

[32] Daojian Zeng, Kang Liu, Siwei Lai, Guangyou Zhou, and Jun Zhao. 2014. Relation Classification via Convolutional Deep Neural Network. In *COLING*. ACL, 2335–2344. https://aclanthology.org/C14-1220/

[33] Yawen Zeng, Qin Jin, Tengfei Bao, and Wenfeng Li. 2023. Multi-Modal Knowledge Hypergraph for Diverse Image Retrieval. In *Thirty-Seventh AAAI Conference on Artificial Intelligence, AAAI 2023, Thirty-Fifth Conference on Innovative Applications of Artificial Intelligence, IAAI 2023, Thirteenth Symposium on Educational Advances in Artificial Intelligence, EAAI 2023, Washington, DC, USA, February 7-14, 2023*, Brian Williams, Yiling Chen, and Jennifer Neville (Eds.). AAAI Press, 3376–3383. https://doi.org/10.1609/AAAI.V37I3.25445

[34] Chuxu Zhang, Dongjin Song, Chao Huang, Ananthram Swami, and Nitesh V Chawla. 2019. Heterogeneous graph neural network. In *Proceedings of the 25th ACM SIGKDD international conference on knowledge discovery & data mining*. 793–803.

[35] Tongtao Zhang, Spencer Whitehead, Hanwang Zhang, Hongzhi Li, Joseph G. Ellis, Lifu Huang, Wei Liu, Heng Ji, and Shih-Fu Chang. 2017. Improving Event Extraction via Multimodal Integration. In *ACM MM*. ACM, 270–278. https://doi.org/10.1145/3123266.3123294

[36] Zizhao Zhang, Haojie Lin, Junjie Zhu, Xibin Zhao, and Yue Gao. 2018. Cross Diffusion on Multi-hypergraph for Multi-modal 3D Object Recognition. In *Advances in Multimedia Information Processing - PCM 2018 - 19th Pacific-Rim Conference on Multimedia, Hefei, China, September 21-22, 2018, Proceedings, Part I (Lecture Notes in Computer Science, Vol. 11164)*, Richang Hong, Wen-Huang Cheng, Toshihiko Yamasaki, Meng Wang, and Chong-Wah Ngo (Eds.). Springer, 38–49. https://doi.org/10.1007/978-3-030-00776-8_4

[37] Changmeng Zheng, Junhao Feng, Ze Fu, Yi Cai, Qing Li, and Tao Wang. 2021. Multimodal Relation Extraction with Efficient Graph Alignment. In *ACM MM*. ACM, 5298–5306. https://doi.org/10.1145/3474085.3476968

[38] Changmeng Zheng, Zhiwei Wu, Junhao Feng, Ze Fu, and Yi Cai. 2021. MNRE: A Challenge Multimodal Dataset for Neural Relation Extraction with Visual Evidence in Social Media Posts. In *ICME*. IEEE, 1–6. https://doi.org/10.1109/ICME51207.2021.9428274

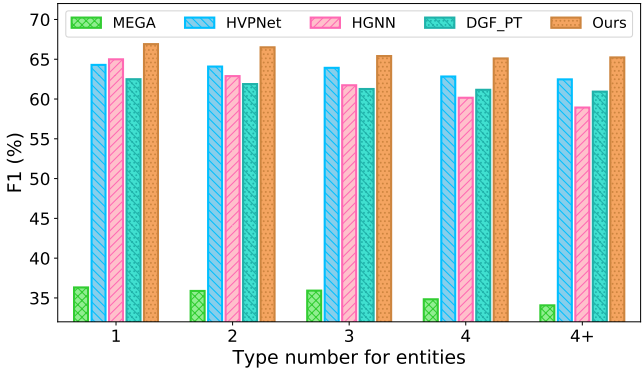

Figure 9: Impact of differences in sample number on MORE. It means the performance (F1) when an entity belongs to one or multiple entity types.

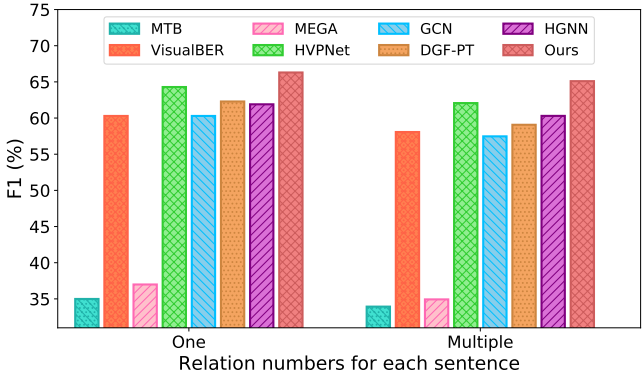

Figure 10: Impact of relation numbers for each sentence on MORE.

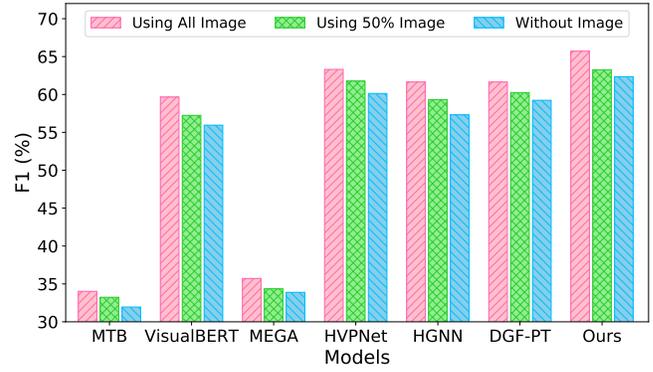

Figure 8: Different proportions of visual information on MORE.

## A  EFFECT OF VISUAL INFORMATION

To further investigate the effectiveness of visual information in our proposed model on other multi-modal relation extraction datasets, we conduct an experiment where we compare the performance of our model with and without image features on the MORE dataset, as shown in Figure 8. From the figure, we can observe that all these observations highlight the importance of capturing high-order correlations between different modalities and considering visual information in MMRE.

## B  DISCUSSIONS FOR V-HAN

To further investigate the impact of multi-semantic entities on relation extraction in our proposed model on other multi-modal relation extraction datasets, we conducted experiments to compare the performance of our proposed method with a baseline method on different numbers of entity repetitions on the MORE dataset, as shown in Figure 9 and 10. The "Without Image" means that all images are deleted. The "Using 50% Image" means that random deletes 50% images from all images. From the figure, we can observe that all these observations highlight the importance of modeling the diversity of multi-semantic entities and utilizing inter-modal associations of different entity pairs to capture the differences on the MMRE task.